\documentclass{article}
\usepackage{spconf,amsmath,graphicx}
\usepackage{balance}
\title{Stereo image de-fencing using Smartphones}

\name{Sankaraganesh Jonna$^1$, Sukla Satapathy$^1$, and Rajiv R. Sahay$^2$}
\address{Computational Vision Laboratory, \\$^1$Department of Computer Science and Engineering, $^2$Department of Electrical Engineering,\\
	$^{1,2}$Indian Institute of Technology Kharagpur}

\begin{document}
%\ninept
%
\maketitle
\begin{abstract}

Conventional approaches to image de-fencing have limited themselves to using only image data in adjacent frames of the captured video of an approximately static scene. In this work, we present a method to harness disparity using a stereo pair of fenced images in order to detect fence pixels. Tourists and amateur photographers commonly carry smartphones/phablets which can be used to capture a short video sequence of the fenced scene. We model the formation of the occluded frames in the captured video.  Furthermore, we propose an optimization framework to estimate the de-fenced image using the total variation prior to regularize the ill-posed problem.

\end{abstract}
\section{Introduction}
\label{sec:int}
	Restoration of images containing occluded objects is a challenging problem and has recently attracted significant attention of researchers. This problem is faced by photographers when they image a scene obstructed by fences e.g. in zoos, museums etc.		
	We observe that image de-fencing is basically an image inpainting \cite{Bertalmio,Criminisi,SP_magazine_2014} problem wherein the fence pixels comprise the region which has to be filled-in. Since fence pixels cover the entire image as shown in Fig. 1 (a), (b) manual segmentation is too cumbersome. In several fence removal methods \cite{Fence_cvpr2008,khasare2013seeing,mu2014video,xue2015computational,yi2016automatic,Jonna_JOSA} a video is captured by panning a camera relative to the scene. Subsequently, these methods use image data in the frames of the video to segment the fences. However, in this work we exploit the disparity of pixels in two adjacent frames to localize fence pixels. Note that generally fences are closer to the panning camera and therefore undergo greater motion parallax due to relative motion.
	
	The main challenge is robust identification of the fences/oc-clusion in bad illumination, heavy clutter in the background, noise etc. In this work, we propose a depth-based technique to estimate the spatial location of fence pixels. Depth maps can be obtained using laser scanners or the inexpensive Kinect sensor. However, laser scanners are slow and expensive and Kinect is limited to capturing depth maps of indoor scenes, respectively. These days smartphones have become ubiquitous and come equipped with sophisticated cameras. Hence, the depth map for estimating fence pixel locations can be obtained using stereo. 
	
\begin{figure}[t]
	\centering
	\begin{tabular}{c c c}
		\includegraphics[width=2.7cm]{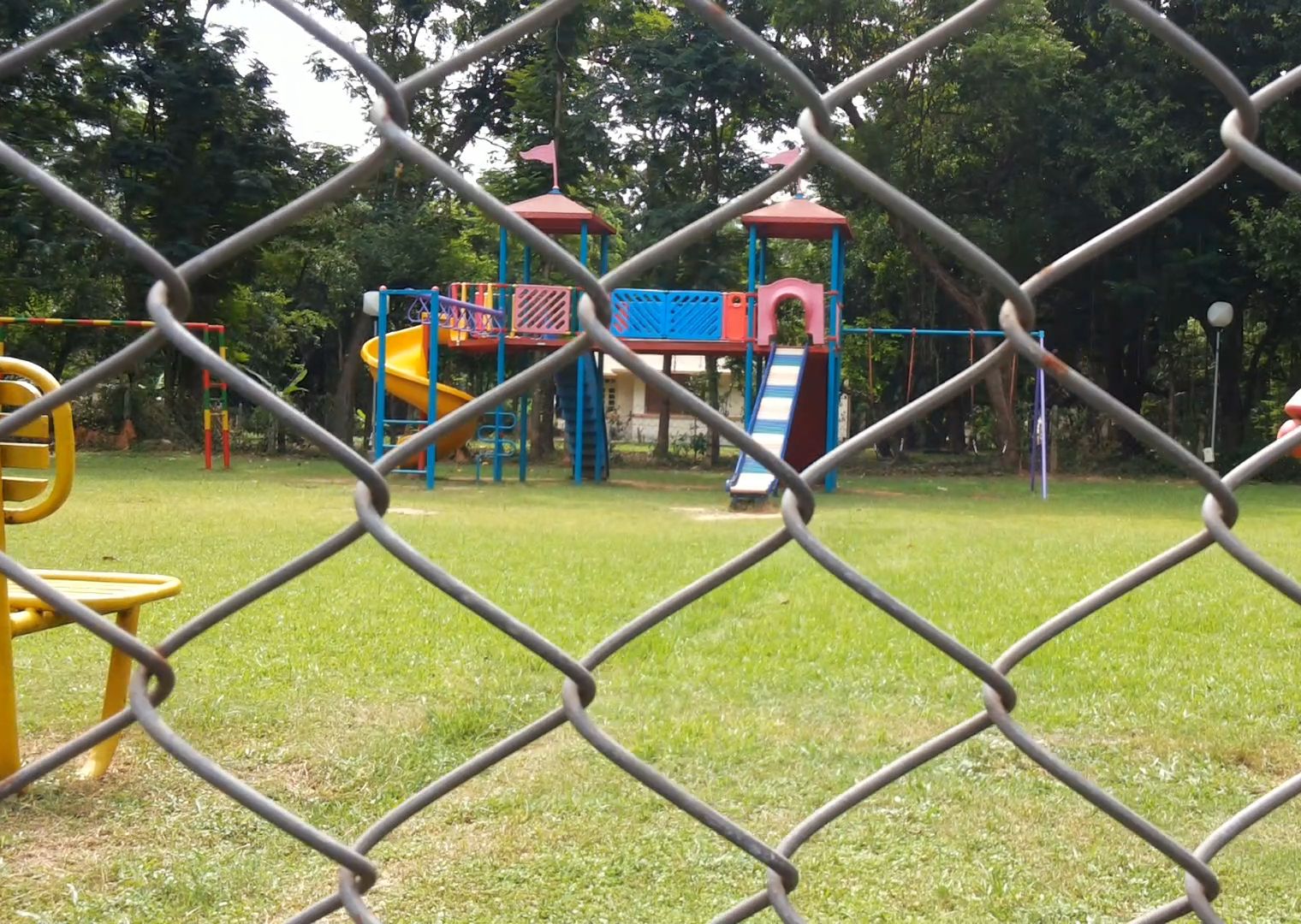}&\hspace*{-12pt}
		\includegraphics[width=2.7cm]{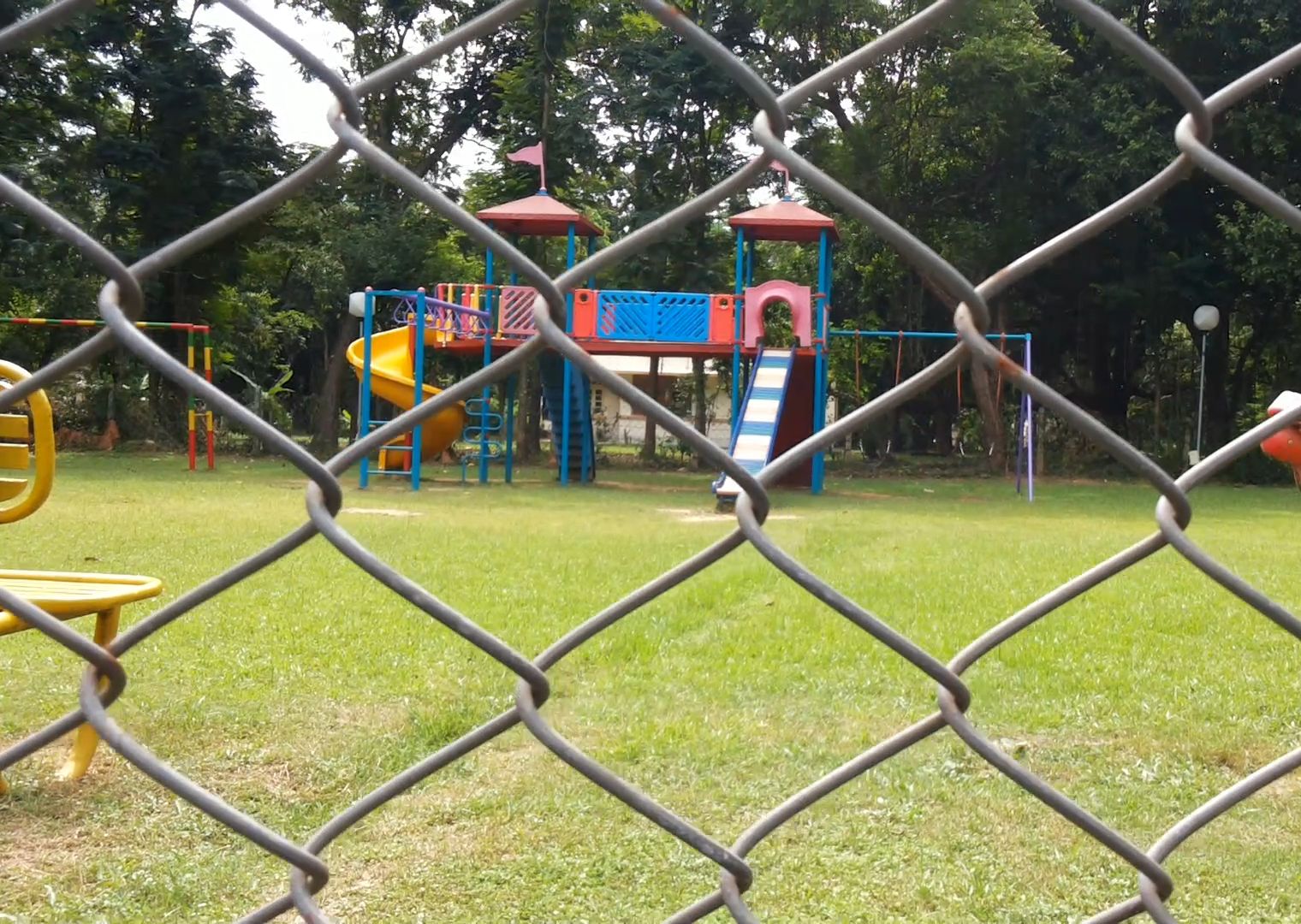}&\hspace*{-12pt}
		\includegraphics[width=2.7cm]{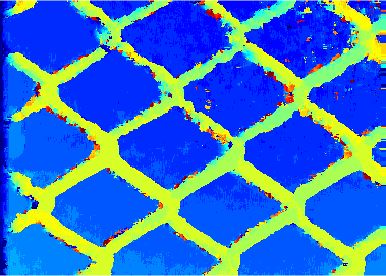}\\
		(a) & (b) & (c)\\
	\end{tabular}
	\caption{(a), (b) Left and right images captured using a mobile phone. (c) Stereo disparity map obtained by proposed algorithm.}
	\label{fig:stereo}
\end{figure}

	In this work, we use only two frames from the captured video using a smartphone and consider them as a stereo-pair to compute the disparity map and subsequently for information fusion. We assume that the objects occluded in left frame are visible in the right image. Zbontar et al. \cite{Lecun_stereo2016} proposed a supervised learning approach to estimate disparity maps by training a convolutional neural network to compare image patches. We employed pre-trained \textquotedblleft fast-CNN" from \cite{Lecun_stereo2016} for estimating a raw disparity map in our work which we refine using morphological filtering and image matting \cite{LBM}. In Figs. \ref{fig:stereo} (a), (b), we show left and right occluded frames captured using a mobile camera. The stereo disparity map computed using the proposed algorithm is shown in Fig. \ref{fig:stereo} (c). In the second step of our algorithm we use the estimated fence pixels to obtain optical flow. Finally, the de-fenced image is reconstructed with total variation (TV) \cite{TV_ROF1992} as the regularizer using the split Bregman \cite{SB_siam2009} optimization framework.

\section{Related work}
\label{sec:format}

Several works have appeared in the literature \cite{Fence_cvpr2008,khasare2013seeing,mu2014video,xue2015computational,yi2016automatic} which use only image data for fence detection. A computational model for periodic pattern perception  in \cite{liu2004computational} is based on the theory of frieze and wallpaper groups.   
Another effective fence mask detection algorithm based on detection of regular textures is proposed by Park et al. in \cite{park2009deformed}. The work in \cite{mu2014video}  addressed the video de-fencing problem using visual parallax as a cue for soft fence  detection distinguishing fences from the background pixels. The occluding foreground along with the restored background are obtained using visual parallax in \cite{xue2015computational}. 
Yi et al. \cite{yi2016automatic} proposed a bottom-up framework for fence detection by clustering pixels into coherent groups using color and motion features through graph-cut based optimization. In our previous works \cite{Jonna_JOSA,Jonna_acpr2015} we have  proposed a supervised learning approach for automatic identification of occlusions/fences using only image data.

In this work,  we propose a depth-based technique to detect the fence mask from the disparity map obtained from a stereo image pair. The works of Wang et al. \cite{stereoInpainting_cvpr08} and Jonna et al. \cite{jonna2015multimodal} are most closely related to our method. The algorithm in \cite{stereoInpainting_cvpr08} leveraged advantages of stereo image pair for joint completion of missing texture and depth. The limitation of \cite{stereoInpainting_cvpr08} is that the region to be inpainted has to be specified manually. The technique of \cite{jonna2015multimodal} proposed a multimodal approach for image de-fencing where the fence masks are extracted automatically from depth maps corresponding to the color images obtained using Kinect sensor. One limitation of the Kinect sensor is that it provides the depth map of a scene in a constrained laboratory environment and only works accurately if the scene is within a particular distance from sensor (viewing distance range is 1.2m to 3.5m). However, most fenced images are captured in outdoor settings such as zoos, play grounds, museums etc. 
The applications of real-world image de-fencing are generally in places, where one can not use the Kinect sensor. Hence, the depth map for estimating fence pixel locations has to be obtained using stereo. 
 
Recently, Zbontar et al. \cite{Lecun_stereo2016} proposed a supervised learning algorithm for estimating disparity map by training convolutional neural network (CNN) architectures \cite{Lecun_1998,ImageNet_NIPS2012}. The raw disparity maps obtained using CNNs in \cite{Lecun_stereo2016} are refined by a series of post-processing steps: cost aggregation, semi-global matching, left-right image consistency check and filtering.
Unlike, the algorithms in \cite{mu2014video,xue2015computational,yi2016automatic}, we use only two frames for both fence segmentation and information fusion. As we are using smartphones for capturing the left and right stereo images we have chosen a learning-based method \cite{Lecun_stereo2016} for disparity estimation wherein we don't require any a priori camera calibration.

\section{Methodology}
\label{sec:pagestyle}

\subsection{ Problem formulation }
\label{sec:typestyle}
The degradation model for the frames of the captured video is
\begin{equation} \label{eq1}
\textbf{y}_m^{obs} = \textbf{O}_{m}\textbf{y}_m = \textbf{O}_m \textbf{W}_{m}\textbf{x} +  \textbf{n}_m 
\end{equation}
where $\textbf{y}_m^{obs}$ is the $m^{\text{th}}$ observation wherein the occluded pixels have been excluded using $\textbf{O}_{m}$, $\textbf{y}_m$, $m={1,2}$ are the left and right images comprising the stereo pair, $\textbf{O}_m$ is the fence mask corresponding to $m^{\text{th}}$ image, $\textbf{W}_m$ is the warp matrix, $\textbf{x}$ is the de-fenced image and $\textbf{n}_m$ is the Gaussian noise.

\subsection{ Detection of fence using stereo} 

In the stereo problem, given a pair of images $\textbf{y}_{m}$ ($m=1,2$) we want to compute the disparity map $\textbf{D}$. Since fences are closer to the camera, we have exploited disparity/depth cue for the fence pattern segmentation.
Following the taxonomy of Scharstein et al. \cite{scharstein2002taxonomy}, a stereo algorithm consists of four steps: matching cost computation, cost aggregation, optmization and disparity refinement. The authors in \cite{Lecun_stereo2016} proposed two supervised learning based convolutional neural network architectures for matching cost computation followed by a set of post processing operations. Note that they trained two CNNs \textquoteleft fast' and \textquoteleft accurate' on pairs of small images patches wherein the true disparity values are known. The advantage of the method in \cite{Lecun_stereo2016} is that it does not require a priori camera calibration. Therefore, we employed the \textquoteleft fast' pre-trained CNN model in \cite{Lecun_stereo2016} which is trained on $17$ million example images extracted from training dataset of KITTI $2015$ \cite{KITTI2015} to generate the raw disparity map.

\begin{figure}[!htb]
	\centering
	\begin{tabular}{c}
		\includegraphics[width=8cm]{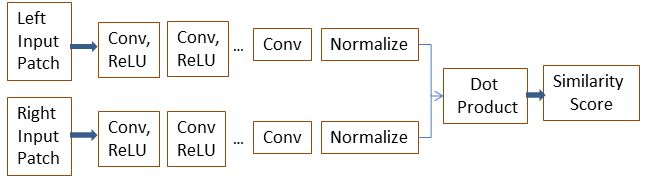}
	\end{tabular}
	\caption{Architecture of the pre-trained \textquoteleft fast' CNN \cite{Lecun_stereo2016}.}
	\label{fig:fastcnn}
\end{figure}
 
The architecture of the \textquoteleft fast' CNN trained for similarity measure on small image patches consists of two sub-networks shown in Fig. \ref{fig:fastcnn}. Each sub-network is made up of convolutional layers followed by ReLU layers except the last convolutional layer. Extracted features from each one of the sub-network are used to compute cosine similarity. The negative value of the cosine similarity measure is treated as the matching cost. To obtain the final fence mask, we use a matting technique \cite{LBM}.

\begin{figure}[!htb]
	\centering
	\begin{tabular}{c}
		\includegraphics[width=8cm]{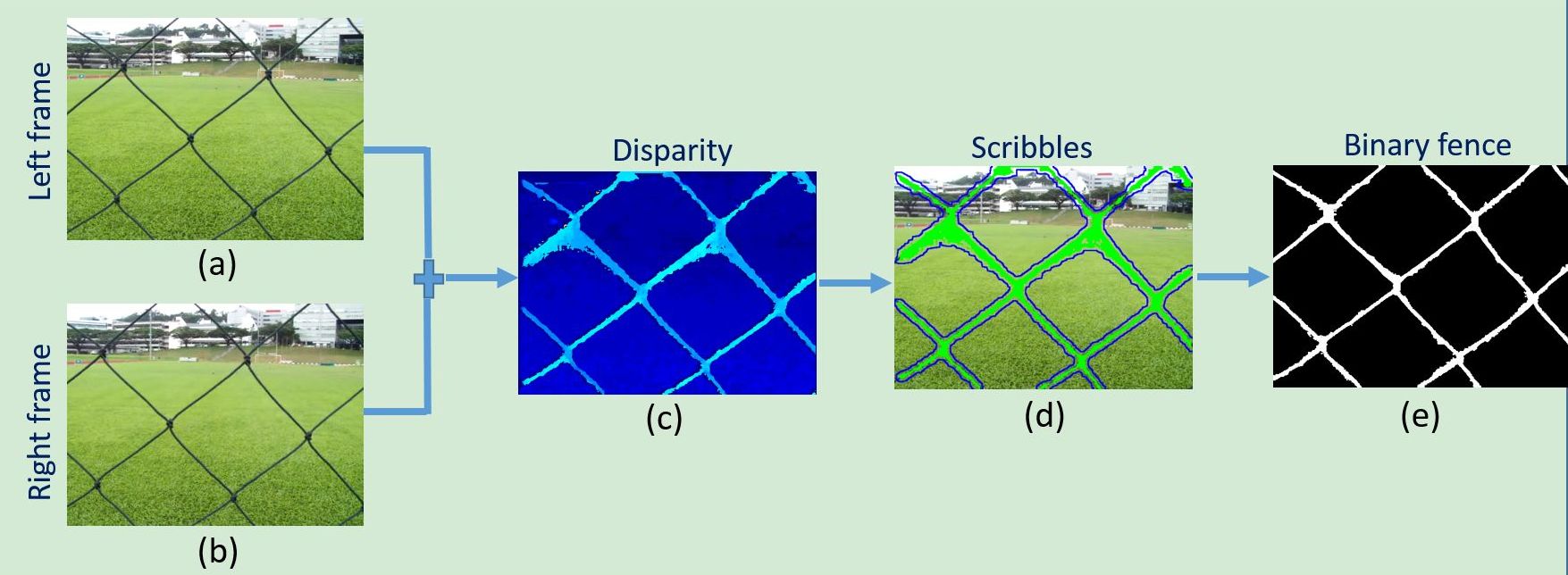}
	\end{tabular}
	\caption{Flowchart of the proposed fence segmentation algorithm.}
	\label{fig:seg_flow}
\end{figure}

The overall workflow of the proposed fence segmentation algorithm is shown in
Fig. \ref{fig:seg_flow}. In Figs. \ref{fig:seg_flow} (a), (b), we show the left and right stereo images taken from a video clip, respectively. Estimated raw disparity map using pre-trained CNN \cite{Lecun_stereo2016} of Fig. \ref{fig:fastcnn} is shown in Fig. \ref{fig:seg_flow} (c). The output in Fig. \ref{fig:seg_flow} (c) is dilated and then canny edge operator is applied. In Fig. \ref{fig:seg_flow} (d) we show the response obtained by Canny edge detection algorithm with blue  color which are treated as background scribbles. Similarly, foreground scribbles (green color) are obtained by erosion operation on the image in Fig. \ref{fig:seg_flow} (c). The combination of both foreground and background scribbles is shown in Fig. \ref{fig:seg_flow} (d). We fed these automatically generated scribbles to the method of \cite{LBM} and obtain the refined binary fence mask shown in Fig. \ref{fig:seg_flow} (e) which is generated by thresholding
the alpha map obtained from \cite{LBM}.

\subsection{Estimation of optical flow} 

The two stereo images in our work have been extracted from a video captured using a smartphone with approximately horizontal motion. To fill-in the occluded information in the reference left frame using visible data from the right view, we need pixel correspondence between them. Optical flow can be estimated accurately between visible areas of two images, however, we observed some erroneous values near the occlusion boundaries. To avoid these errors, we blurred the fences in the stereo images with a Gaussian kernel ($\sigma=2$) before applying the method in \cite{Flow_Brox20111} to estimate optical flow. Note that now we already know the spatial locations of the fence pixels which are estimated in the previous section.

\subsection{ Optimization using TV} 
As we model image de-fencing as an inverse ill-posed problem, total variation of the de-fenced image can be posed as a regularization constraint. Since TV prior is non-smooth split Bregman technique \cite{SB_siam2009} is used to solve the resulting minimization problem to reconstruct the fence-free image.
The de-fenced image is the solution of the following constrained optimization problem
\begin{figure}[!htb]
	\centering
	\begin{tabular}{c c c}
		\includegraphics[width=2.5cm]{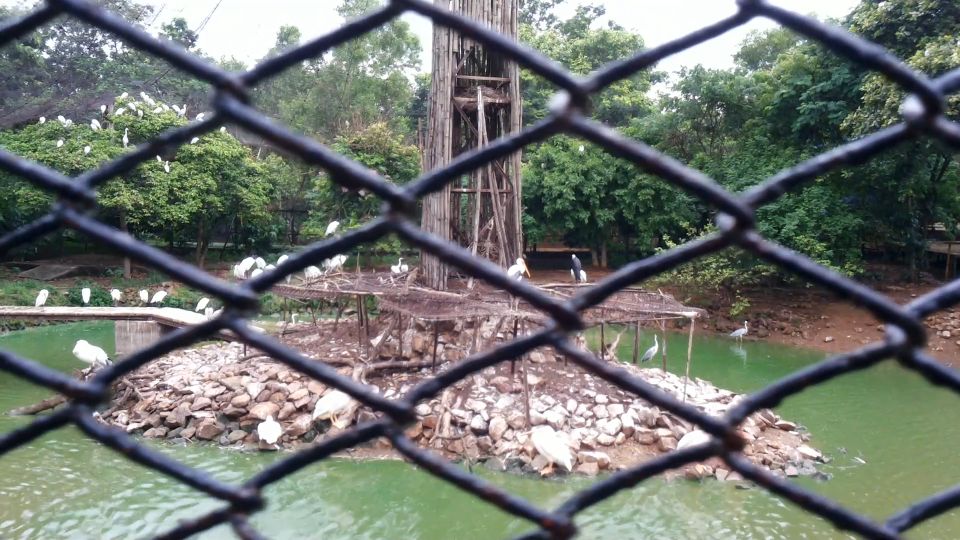}&\hspace*{-11pt}
		\includegraphics[width=2.5cm]{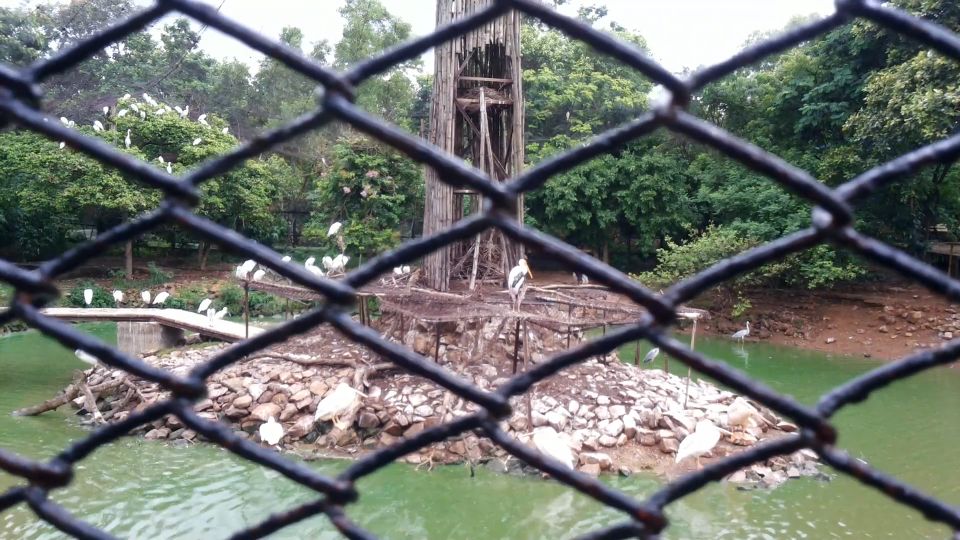}&\hspace*{-11pt}
		\includegraphics[width=2.5cm]{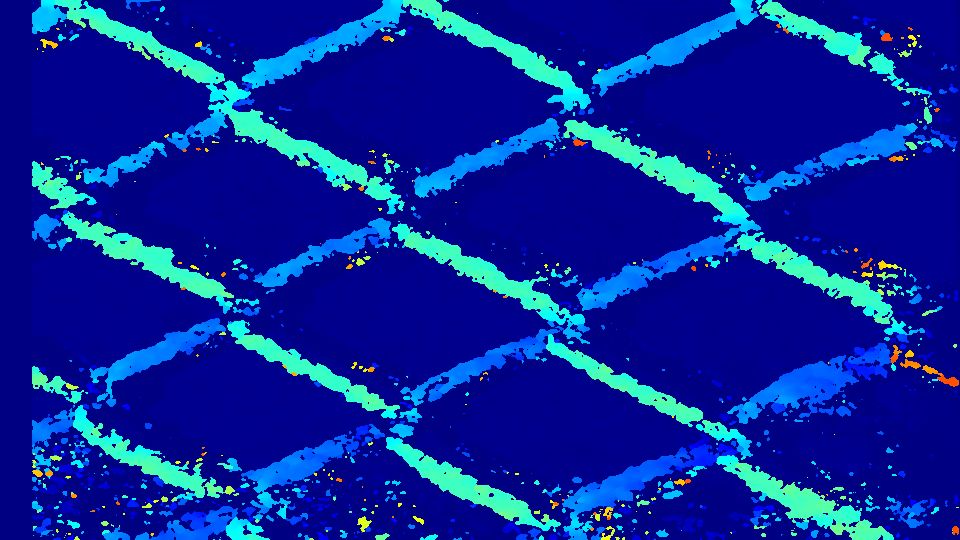}\\
		(a) & (b) & (c)\\
		\includegraphics[width=2.5cm]{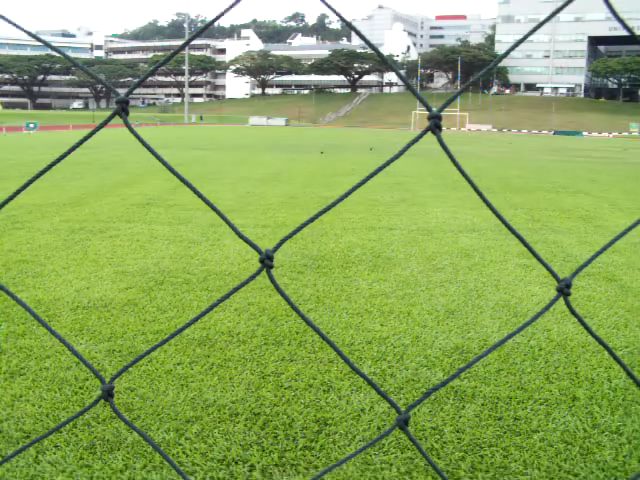}&\hspace*{-12pt}
		\includegraphics[width=2.5cm]{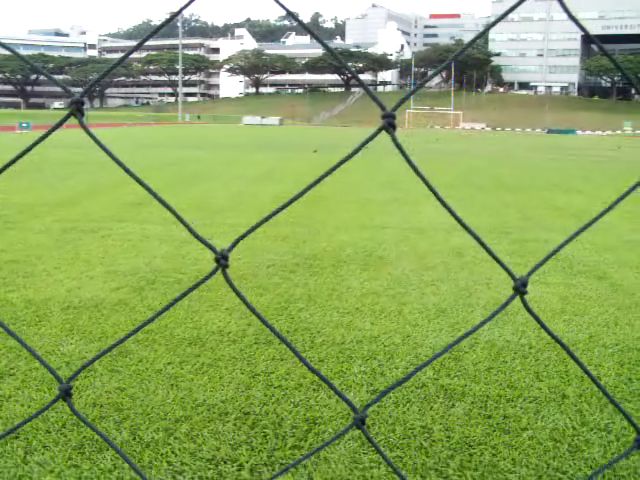}&\hspace*{-12pt}
		\includegraphics[width=2.5cm]{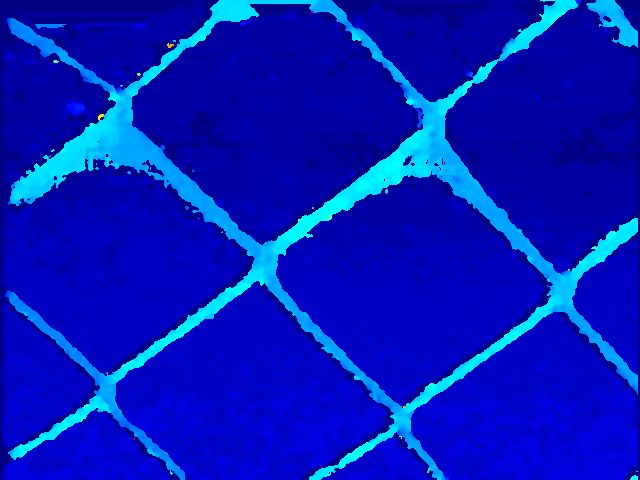}\\
		(d) & (e) & (f)\\
		\includegraphics[width=2.5cm]{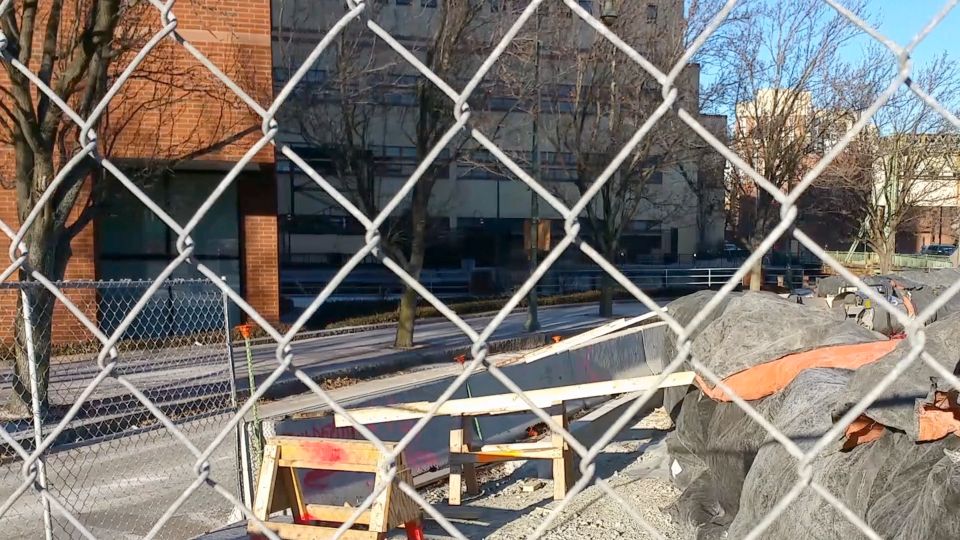}&\hspace*{-12pt}
		\includegraphics[width=2.5cm]{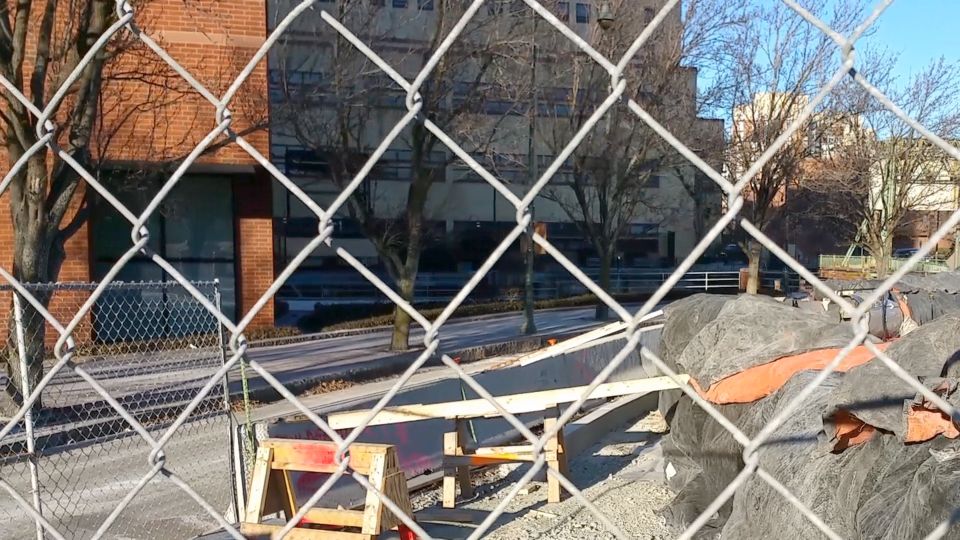}&\hspace*{-12pt}
		\includegraphics[width=2.5cm]{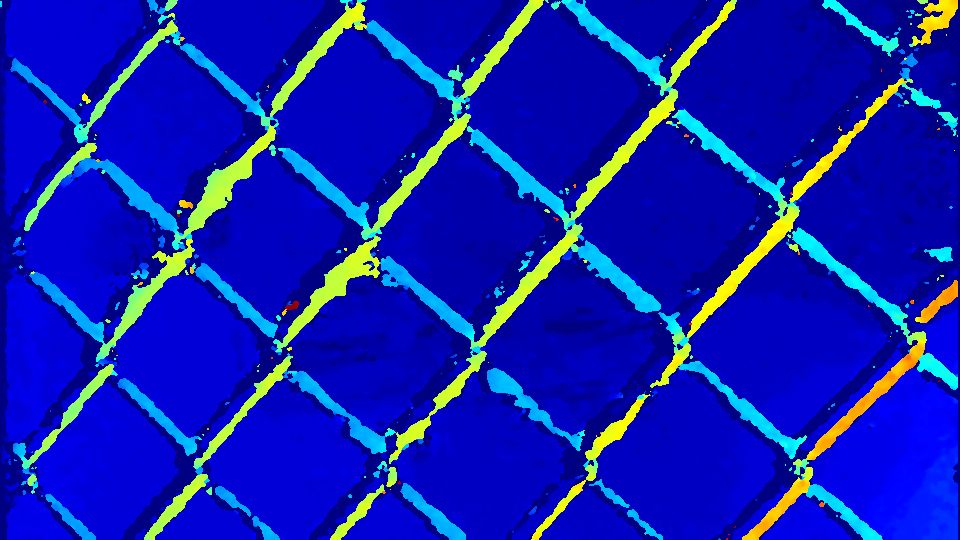}\\
		(g) & (h) & (i)\\-
		\includegraphics[width=2.5cm]{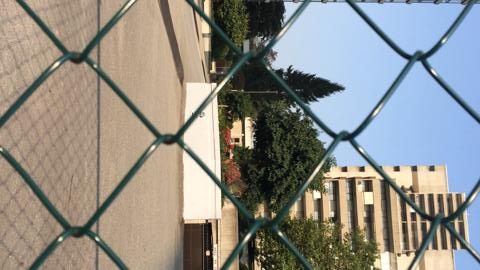}&\hspace*{-11pt}
		\includegraphics[width=2.5cm]{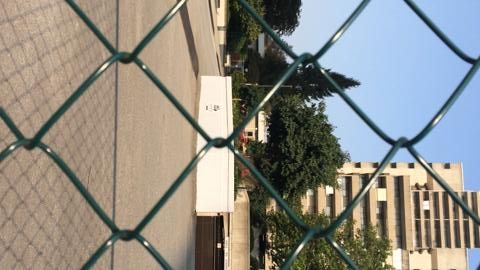}&\hspace*{-11pt}
		\includegraphics[width=2.5cm]{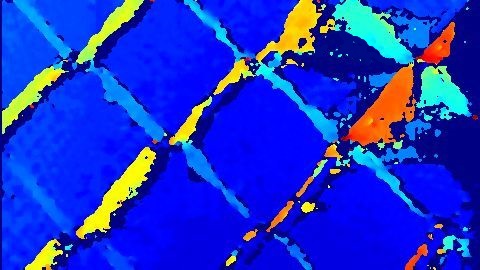}\\
		(j) & (k) & (l)\\
	\end{tabular}
	\caption{Row 1: First and second columns show the left and right stereo frames captured by us using a smartphone. Rows $2$ to $4$: First and second columns depict stereo pairs from video clips of \cite{mu2014video,xue2015computational,yi2016automatic}, respectively. Third column: estimated disparity maps using the proposed algorithm.}
\end{figure}
\begin{equation*}
\small
\arg\min_{\textbf{x}}  \frac{1}{2} \sum_{m=1}^{p}\parallel  \textbf{y}_{m}-\textbf{O}_{m}\textbf{W}_{m}\textbf{x}\parallel _{2}^{2} +  \mu \parallel  \textbf{d} \parallel_{1}  s.t. \hspace{5pt}\textbf{d} = \nabla   \textbf{x} 
\end{equation*}
where $p$ is the number of frames chosen from the video and $\mu$ is the regularization parameter. The above optimization framework is a combination of both $l1$ and $l2$ terms and hence difficult to solve. We employ the split Bregman iterative framework described in \cite{SB_siam2009} to solve the above problem. We use an alternative unconstrained formulation as
\begin{equation}
\begin{split}
\small
\arg\min_{\textbf{x}}  \frac{1}{2} \sum_{m=1}^{p}\parallel  \textbf{y}_{m}-\textbf{O}_{m}\textbf{W}_{m}\textbf{x}\parallel _{2}^{2}  + \mu \parallel  \textbf{d} \parallel_{1} \\
+\frac{ \lambda }{2}  \parallel \textbf{d}- \nabla \textbf{x} \parallel_{2}^{2} 
\end{split}
\end{equation}
where $\lambda$ is the shrinkage parameter.
The iterates to solve the above equation are as
\begin{equation}
\begin{split}
\small
[\textbf{x}^{k+1},\textbf{d}^{k+1}] = \arg\min_{\textbf{x},\textbf{d}} \ \frac{1}{2} \sum_{m=1}^{p}\parallel  \textbf{y}_{m}-\textbf{O}_{m}\textbf{W}_{m}\textbf{x}^{k}\parallel _{2}^{2} \\
+ \mu \parallel  \textbf{d}^{k} \parallel_{1}
+\frac{ \lambda }{2}  \parallel \textbf{d}^{k}- \nabla \textbf{x}^{k} + \textbf{b}^{k}\parallel_{2}^{2}
\end{split}
\end{equation}
\begin{equation*}
\small
\textbf{b}^{k+1} = \nabla \textbf{x}^{k+1} + \textbf{b}^{k} -\textbf{d}^{k+1}
\end{equation*}
We can now split the above problem into two sub-problems as \\
\textbf{Sub Problem 1:}
\begin{equation*}	
\begin{split}
\small
[\textbf{x}^{k+1}] = \arg\min_{\textbf{x}} \ \frac{1}{2} \sum_{m=1}^{p}\parallel  \textbf{y}_{m}-\textbf{O}_{m}\textbf{W}_{m}\textbf{x}^{k}\parallel _{2}^{2}  \\
+ \frac{ \lambda }{2}  \parallel \textbf{d}^{k}- \nabla \textbf{x}^{k} + \textbf{b}^{k}\parallel_{2}^{2}
\end{split}
\end{equation*}
This sub-problem is solved by a steepest descent method.\\
\textbf{Sub Problem 2:} 
\begin{equation*}	
\small
[\textbf{d}^{k+1}] = \arg\min_{\textbf{d}} \mu \parallel  \textbf{d}^{k} \parallel_{1} + \frac{ \lambda }{2}  \parallel \textbf{d}^{k}- \nabla \textbf{x}^{k+1} + \textbf{b}^{k}\parallel_{2}^{2}
\end{equation*}
The above sub-problem can be solved by applying the shrinkage operator as follows
\begin{equation*}
\small
\textbf{d}^{k+1}=shrink(\nabla \textbf{x}^{k+1}+\textbf{b}^{k},\frac{\lambda}{\mu})
\end{equation*}
\begin{equation*}
\begin{split}
\small
\textbf{d}^{k+1} = \frac{\nabla \textbf{x}^{k+1}+\textbf{b}^{k}}{\mid \nabla \textbf{x}^{k+1}+\textbf{b}^{k} \mid}*max(\mid \nabla \textbf{x}^{k+1}+\textbf{b}^{k} \mid - \frac{\lambda}{\mu},0) 
\end{split}
\end{equation*}
The variable \textbf{b} can be updated as 
$ \textbf{b}^{k+1} = \nabla \textbf{x}^{k+1} + \textbf{b}^{k} - \textbf{d}^{k+1}$.
To obtain the best estimate of the de-fenced image we tuned the parameters $\mu$, $\lambda$.

\section{Experimental results}
\label{sec:majhead}

To validate the robustness of the proposed method we have reported both fence detection and image de-fencing results using videos captured by us with a smartphone as well as video sequences reported in \cite{mu2014video,xue2015computational,yi2016automatic}. We have generated all the results using a $3.4$ GHz Intel Core i$7$ processor with $16$ GB of RAM. The execution time of our non-optimized MATLAB implementation is of the order of few tens of seconds.

Initially, we report the fence segmentation results. The images shown in Figs. 4 (a) and (b) are taken from a short video clip captured by us using a smartphone camera. In Fig. 4 (c), we show the raw disparity map obtained using the pre-trained CNN \cite{Lecun_stereo2016} of Fig. \ref{fig:fastcnn} which is post-processed by morphological operations and learning-based matting method of \cite{LBM} as outlined in section $3.2$. Next, we conduct an experiment on a dataset taken from \cite{mu2014video}. In Figs. 4 (d) and (e), we show the stereo pair and the computed disparity map is depicted in Fig. 4 (f). We conducted one more experiment using a video reported in \cite{xue2015computational}. In Figs. 4 (g) and (h) we show two frames taken from their video sequence and the corresponding disparity map estimated using our algorithm is shown in Fig. 4 (i). Finally, we used two frames from a video sequence reported in  \cite{yi2016automatic} which are shown in Figs. 4 (j) and (k). In Fig. 4 (l) we show the obtained disparity map using our approach.

\begin{figure}[!htb]
	\centering
	\begin{tabular}{c c c}
		\includegraphics[width=2.5cm]{BBS129_c1.jpg}&\hspace*{-11pt}
		\includegraphics[width=2.5cm]{BBS129_c4.jpg}&\hspace*{-11pt}
		\includegraphics[width=2.5cm]{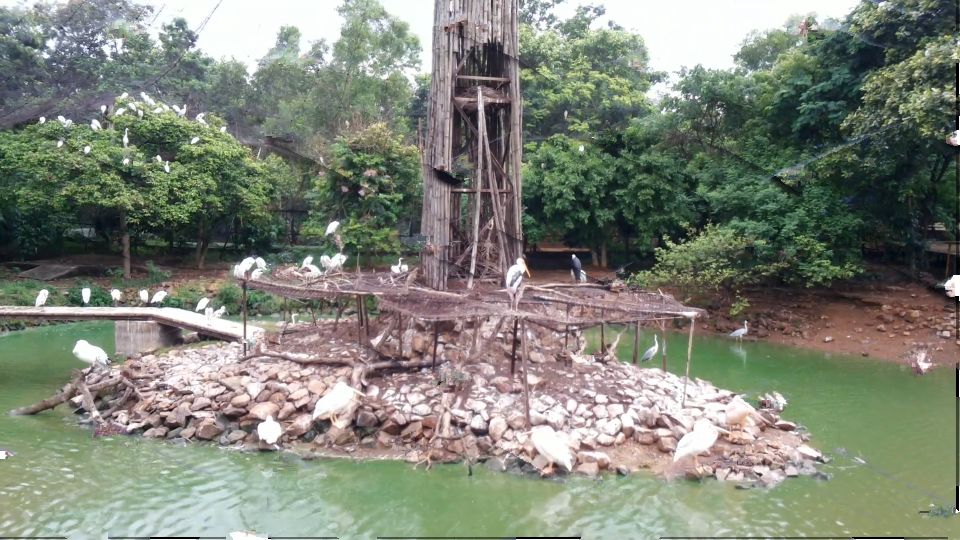}\\
		(a) & (b) & (c)\\
		\includegraphics[width=2.5cm]{tcsvt_1.jpg}&\hspace*{-11pt}
		\includegraphics[width=2.5cm]{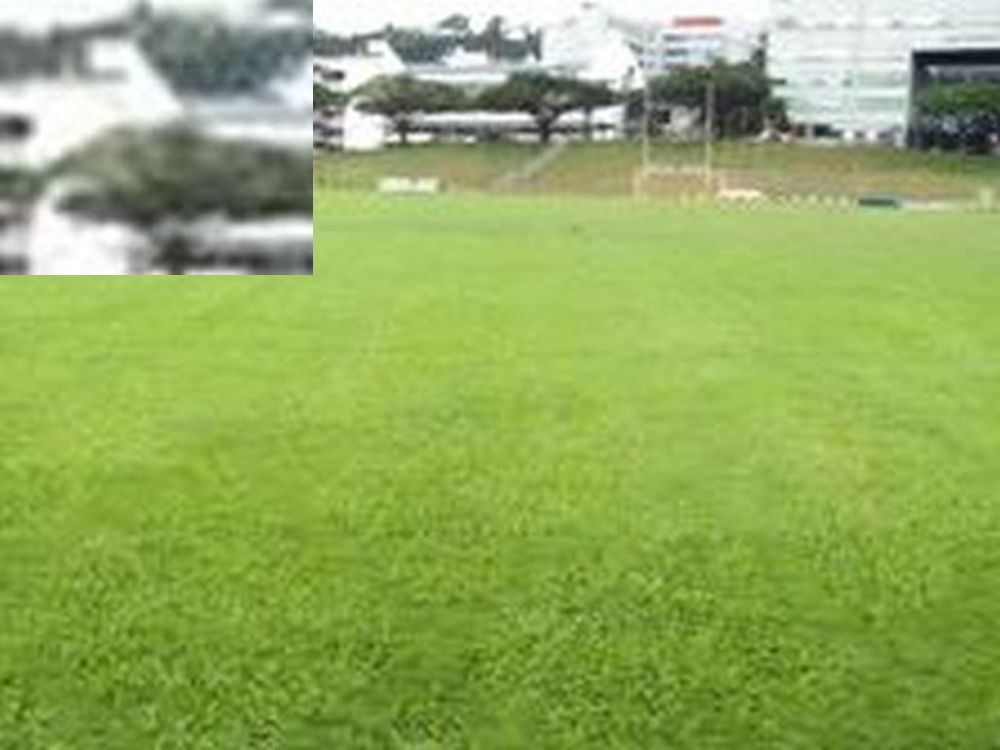}&\hspace*{-11pt}
		\includegraphics[width=2.5cm]{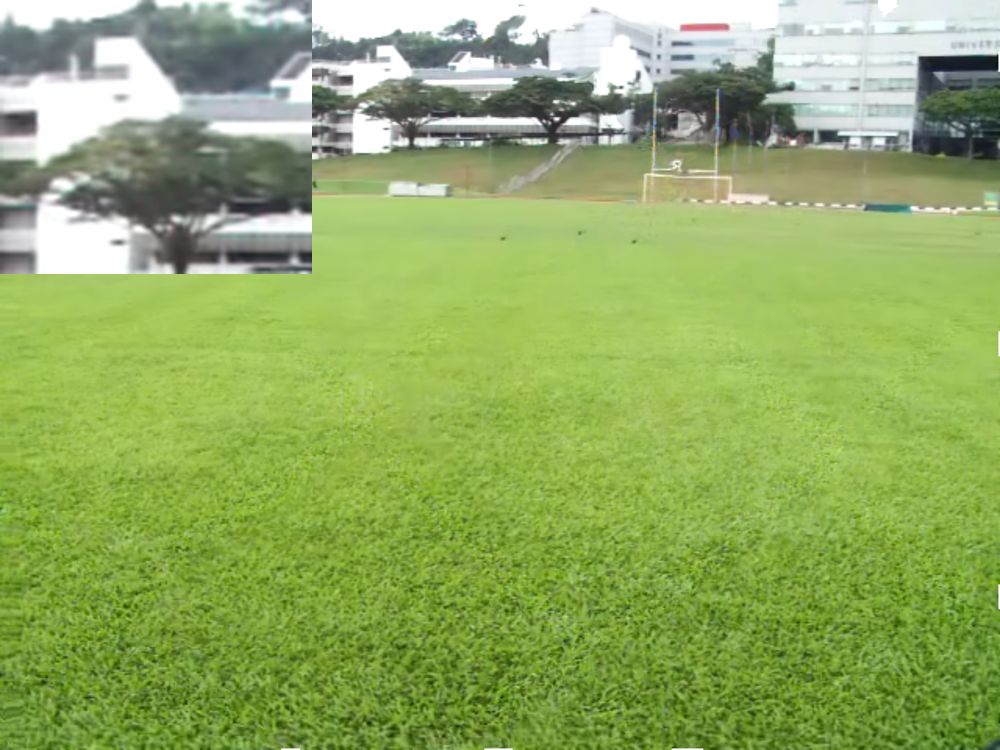}\\
		(d) & (e) & (f)\\
		\includegraphics[width=2.5cm,height=1.4cm]{tog_c1.jpg}&\hspace*{-11pt}
		\includegraphics[width=2.5cm,height=1.4cm]{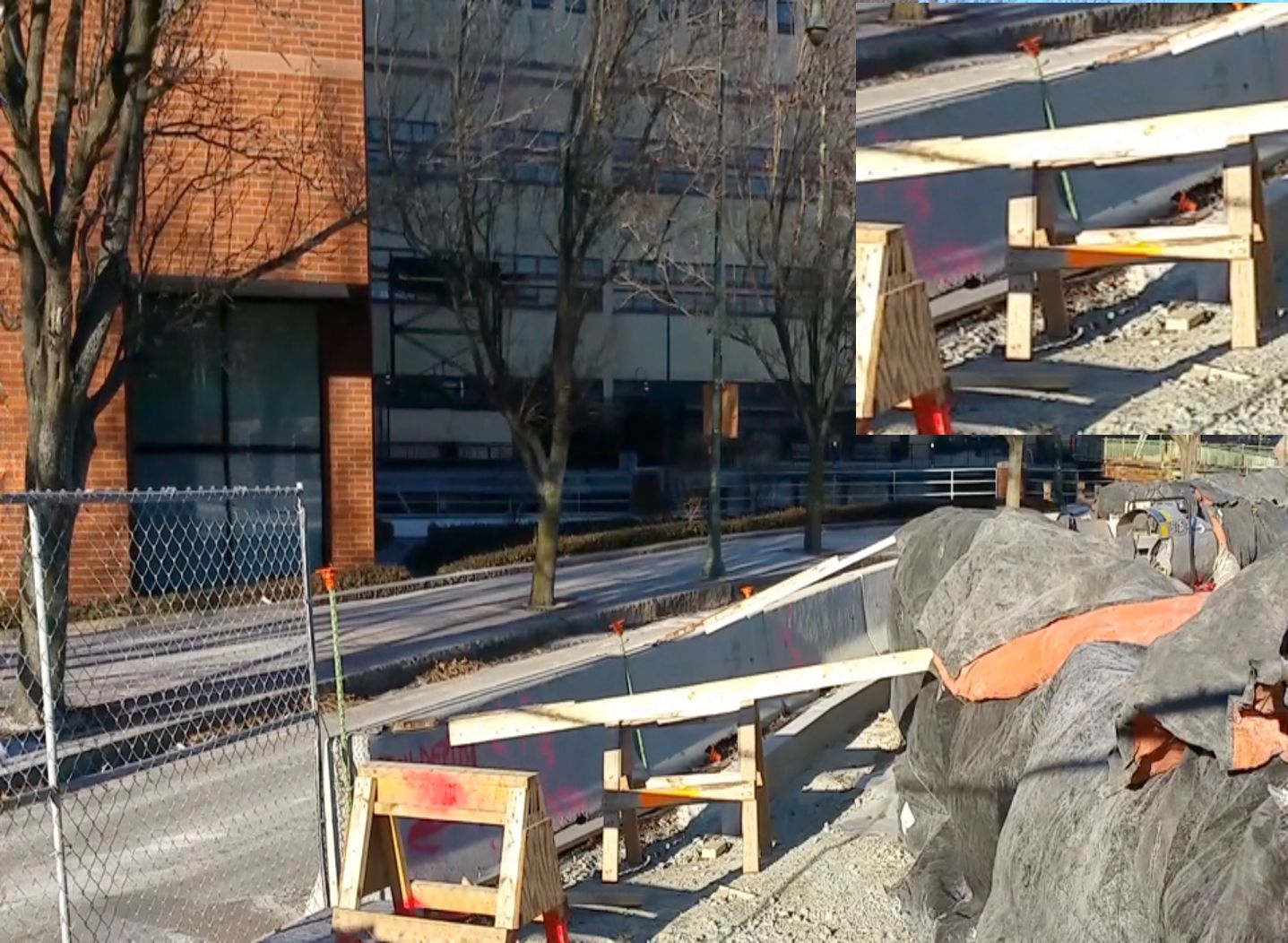}&\hspace*{-11pt}
		\includegraphics[width=2.5cm,height=1.4cm]{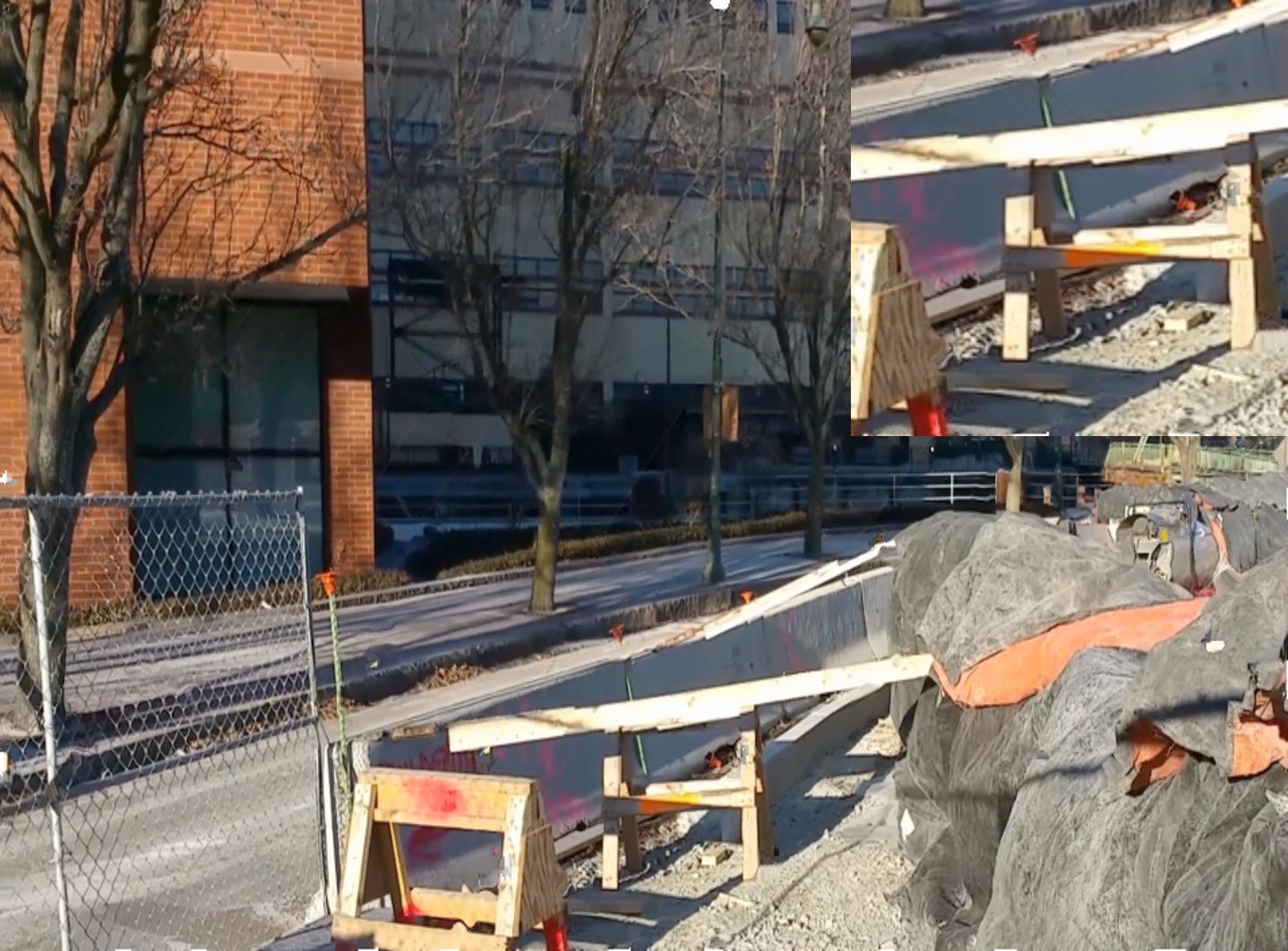}\\
		(g) & (h) & (i)\\
		\includegraphics[width=2.5cm,height=1.4cm]{cvpr_c2.jpg}&\hspace*{-11pt}
		\includegraphics[width=2.5cm,height=1.4cm]{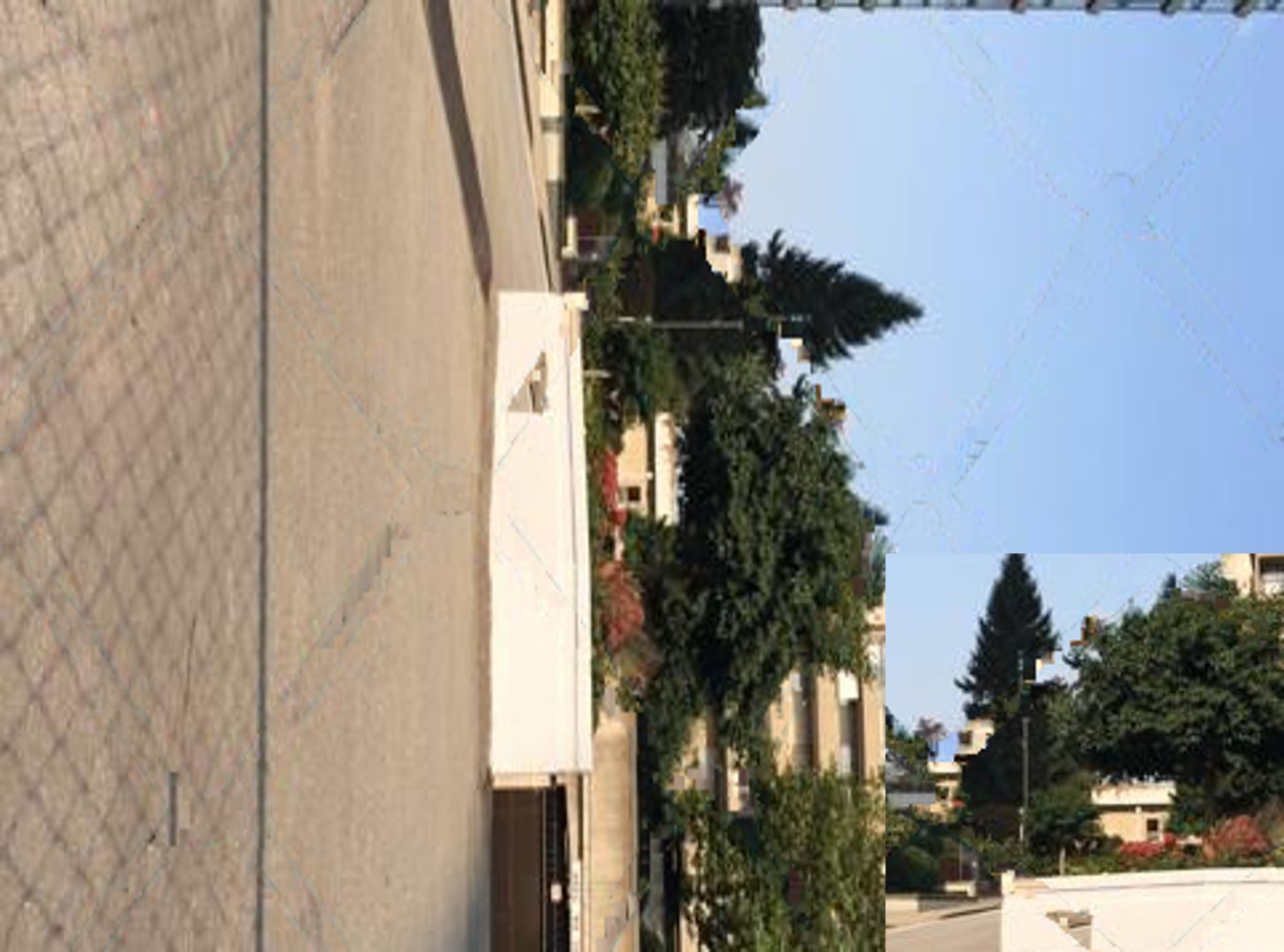}&\hspace*{-11pt}
		\includegraphics[width=2.5cm,height=1.4cm]{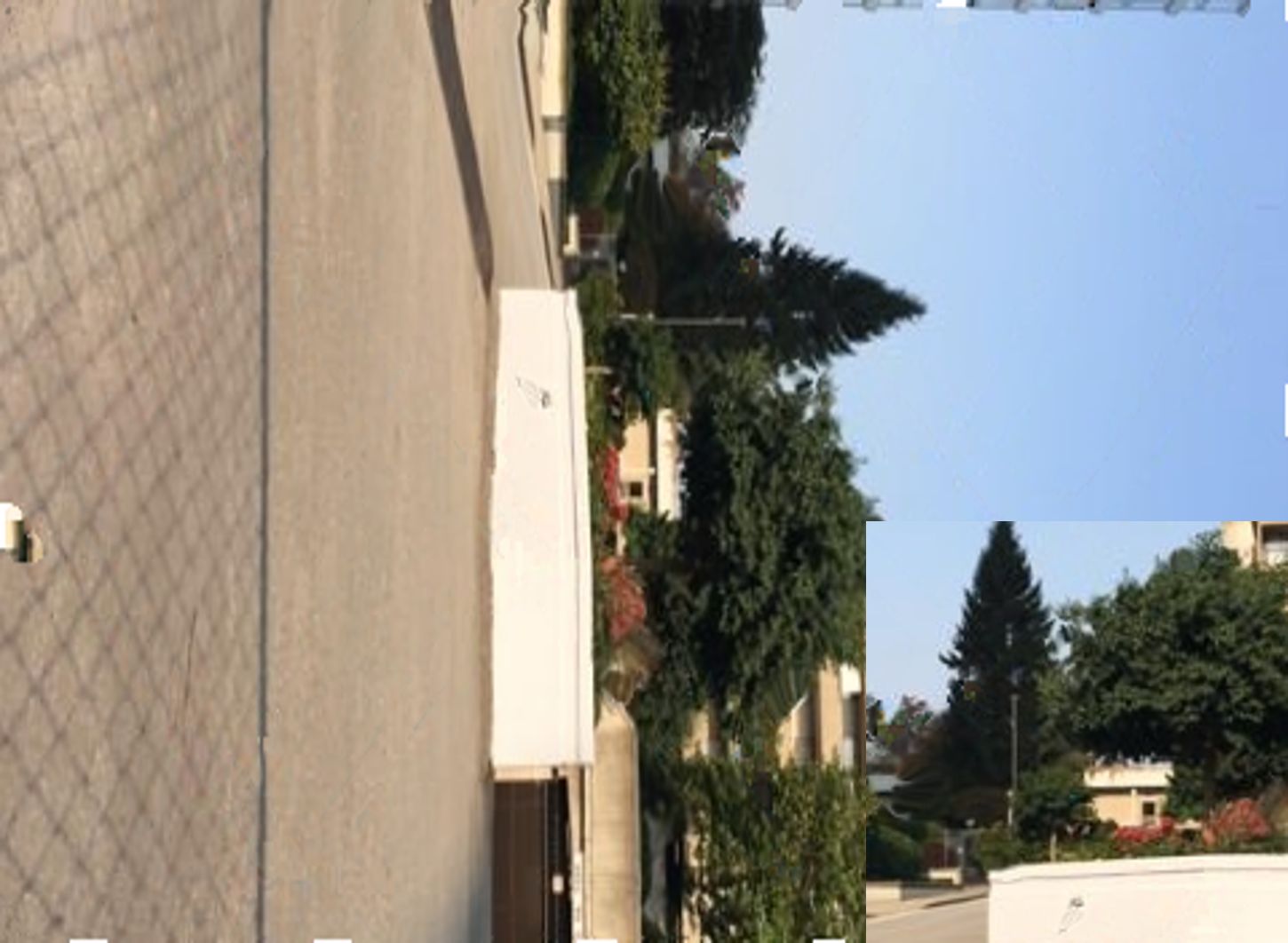}\\
		(j) & (k) & (l)\\
	\end{tabular}
	\caption{(a), (b) Two fenced observations from a video captured by us using a smartphone. (c) De-fenced image obtained by the proposed technique. Rows $2$ to $4$: Comparisons with the existing image/video de-fencing algorithms. \textbf{First column}: Sample reference images from video sequences \cite{mu2014video,xue2015computational,yi2016automatic}, respectively. \textbf{Second column}: De-fenced images obtained using state-of-the-art image de-fencing methods \cite{mu2014video,xue2015computational,yi2016automatic}, respectively. \textbf{Third column}: De-fenced results obtained using the proposed algorithm.}% corresponding to images shown in the first column.}
\end{figure}

Now we show the performance of the proposed algorithm for removing fences. In Fig. 5 (a) and (b) we show two fenced images taken from a video sequence captured by us using a smartphone. The de-fenced image is shown in Fig. 5 (c). Note that there are no residual traces of fences in the reconstructed output wherein all details have been preserved.
We now provide comparisons with state-of-the-art image/video de-fencing algorithms using video sequences reported from their works \cite{mu2014video,xue2015computational,yi2016automatic}. 
In Fig. 5 (d), we show the reference frame taken from a video sequence in \cite{mu2014video}. The de-fenced images obtained using the method of \cite{mu2014video} and the proposed algorithm are shown in Figs. 5 (e) and (f), respectively. Note that de-fenced image obtained using \cite{mu2014video} is blurred whereas the proposed algorithm preserves sharp details. This is evident in the close-up insets of Figs. 5 (e) and (f). Next, a frame taken from a video sequence reported in \cite{xue2015computational} is shown in Fig. 5 (g). In Figs. 5 (h) and (i), we show the de-fenced results obtained by \cite{xue2015computational} and the proposed algorithm, respectively. Interestingly, using only $2$ frames the proposed algorithm achieves comparable results to \cite{xue2015computational} which in contrast used $5$ frames. Finally, we perform an experiment on a recent video sequence reported in \cite{yi2016automatic}. A frame taken from the video is shown in Fig. 5 (j). The inpained image obtained using \cite{yi2016automatic} is shown in Fig. 5 (k). The corresponding de-fenced image obtained using the proposed algorithm is shown in Fig. 5 (l). Observe that the result in Fig. 5 (k) contains some artifacts whereas the obtained de-fenced image in Fig. 5 (l) is reconstructed well. The same can be observed in insets of both the Figs. 5 (k) and (l), respectively. Note that the algorithm in \cite{yi2016automatic} employed inpainting technique of \cite{Criminisi} for image de-fencing.

\section{Conclusions}
In this work, we present a novel algorithm for fence segmentation and removal using a stereo-pair of fenced images captured with a smartphone. Initially, we harnessed the disparity cue for robust fence segmentation. We computed the raw disparity map corresponding to a pair of images using pre-trained CNNs \cite{Lecun_stereo2016} followed by a series of post-processing steps for obtaining an accurate fence mask. Subsequently, an optimization framework is formulated and the ill-posed inverse problem of image de-fencing is solved using split Bregman iterative procedure assuming total variation of the de-fenced image as the regularization constraint. We compare the performance of the proposed algorithm with several state-of-the-art works for image de-fencing.

%\vfill\pagebreak

\balance

\bibliographystyle{ieeetran}
\bibliography{icassp_db2}
%\bibliography{strings,ICASSP_ref1}
%\begin{thebibliography}

\end{document}